\def\year{2016}\relax
\documentclass[letterpaper]{article}
\usepackage{aaai17}
\usepackage{times}
\usepackage{helvet}
\usepackage{courier}

\newcommand{\cited}[1]{\citeauthor{#1}~\shortcite{#1}}

\usepackage{amsmath,amssymb,amsfonts,amsthm}
\usepackage{bbm}
\usepackage{mathrsfs}

\usepackage{graphicx}

\usepackage{algorithm}
\usepackage{algpseudocode}

\theoremstyle{plain}
\newtheorem{thm}{Theorem}

\newtheorem{prop}[thm]{Proposition}

\theoremstyle{definition}

\theoremstyle{remark}

\algrenewcommand{\algorithmicrequire}{\textbf{Input:}}
\algrenewcommand{\algorithmicensure}{\textbf{Output:}}

\renewcommand\[{\begin{equation}}
\renewcommand\]{\end{equation}}

\let\emptyset\varnothing

\newcommand{\bbR}{\mathbb{R}}

\newcommand{\calvar}[1]{\ensuremath{\mathcal{#1}}}
\newcommand{\calD}{\calvar{D}}
\newcommand{\calX}{\calvar{X}}

\newcommand{\vecvar}[1]{\ensuremath{\boldsymbol{#1}}}
\newcommand{\va}{\vecvar{a}}
\newcommand{\vb}{\vecvar{b}}

\newcommand{\vw}{\vecvar{w}}
\newcommand{\vz}{\vecvar{z}}

\newcommand{\vphi}{\vecvar{\phi}}
\newcommand{\vepsilon}{\vecvar{\varepsilon}}

\DeclareMathOperator*{\argmax}{argmax}
\DeclareMathOperator*{\argmin}{argmin}

\usepackage{color}

\newcommand{\cc}{\textsc{cc}}
\newcommand{\cl}{\textsc{cl}}

\begin{document}
\title{Coactive Critiquing: Elicitation of Preferences and Features}
\author{Stefano Teso\\
\texttt{stefano.teso@unitn.it}\\
University of Trento\\
Via Sommarive 9, Povo\\
Trento, Italy\\
\And
Paolo Dragone\\
\texttt{paolo.dragone@unitn.it}\\
University of Trento\\
TIM-SKIL\\
Via Sommarive 9, Povo\\
Trento, Italy\\
\And
Andrea Passerini\\
\texttt{andrea.passerini@unitn.it}\\
University of Trento\\
Via Sommarive 9, Povo\\
Trento, Italy\\
}
\maketitle

\begin{abstract}
When faced with complex choices, users refine their own preference criteria as
they explore the catalogue of options. In this paper we propose an approach to
preference elicitation suited for this scenario. We extend Coactive Learning,
which iteratively collects manipulative feedback, to optionally query example
critiques. User critiques are integrated into the learning model by dynamically
extending the feature space. Our formulation natively supports constructive
learning tasks, where the option catalogue is generated on-the-fly. We
present an upper bound on the average regret suffered by the learner. Our
empirical analysis highlights the promise of our approach.
\end{abstract}

\section{Introduction}

Preference elicitation~\cite{goldsmith2009preference} is the task of
interactively inferring preferences of users and it is a key component of
personalized recommendation and decision support systems. The typical approach
consists of asking the user to rank alternative
solutions~\cite{chajewska2000,boutilier2006,guo2010real,viappiani2010optimal}
and use the resulting feedback to learn a (possibly approximately) consistent
user utility function. These algorithms rely on a fixed pool of solutions from
which to choose both candidates for feedback and final recommendations.
However, when thinking of an interaction between a user and a salesman, one
imagines a more active role by the user, who could suggest
modifications to candidates. For instance, in a trip planning application, when
commenting a candidate trip to New York, the user may reply: ``I'd rather visit the
MoMA than Central Park''. This is
especially true when considering fully {\em constructive}
scenarios~\cite{teso2016constructive}, where the task is synthesizing entirely
novel objects, like the furniture arrangement of an apartment or a novel recipe
for vegan tiramis\`u. Coactive Learning~\cite{shivaswamy2012online} is a recent
interactive learning paradigm which allows the user to provide corrected
versions of the candidates she is presented with.

While Coactive Learning approaches adapt the preference model based on
user-provided option improvements, the set of features that the utility is
defined by is assumed given and fixed. This is not always a realistic
assumption. When faced with a complex decision, users may not be fully aware of
their own quality criteria, especially in large, unfamiliar decision
domains~\cite{chen2012critiquing,pu2000enriching}. Even more so in constructive
settings, where the option catalogue is exponentially (possibly infinitely)
large and generated on-the-fly. Crucially, the user may become aware of novel
preference criteria, in a context-specific fashion, while exploring the decision
domain~\cite{payne1993adaptive,slovic1995construction}.

One way to tackle this problem is to enumerate all potential user criteria in
advance, by combining a fixed set of features with one or more operators (e.g.
multiplication or logical conjunction). This solution however has drawbacks.
First, the number of feature combinations suffers from combinatorial explosion,
making learning harder and more computationally demanding. Most importantly,
entirely novel and unanticipated user criteria can not be added to the feature
space.

Example critiquing (or conversational) recommendation
systems~\cite{tou1982rabbit,mcginty2011evolution,chen2012critiquing} provide an
alternative solution. In this setting, preferences are stated in term of
critiques to suggested configurations. Upon receiving one or more proposals,
the user is free to reply with statements such as ``this trip is too
expensive'' or ``I dislike crowded places''. Critiques are integrated into
the learner as auxiliary constraints or penalties~\cite{faltings2004designing}.  Options
presented at later iterations are chosen based on the collected feedback,
focusing the search on more promising items. Example critiquing is explicitly
designed to address the above difficulty: by being confronted with a set of
concrete items, the user has a chance to realize that she cares about features
that she was previously unaware of~\cite{chen2012critiquing}. Unfortunately,
typical conversational systems do not support numerical modelling of user
preferences (e.g. weights), and often assume noiseless critiquing feedback.

In this paper we present a new algorithm, Coactive Critiquing (\cc), that unifies
\textit{coactive} learning and example \textit{critiquing}, harnessing the
strengths of both strategies. Coactive Critiquing builds on the coactive
learning framework by further allowing critique feedback. We view critiques as
arbitrarily articulated explanations for the user-provided improvements, e.g.
the user may explain her reason for suggesting the MoMA over Central Park
by stating: ``I prefer indoor activities during winter''.  In this work, we
assume that there is an interface between the algorithm and the user which
translates the user's critiques into (soft) constraints\footnote{For instance,
it could be a simple form that allows the user to combine attribute values to
form critiques. We are currently working on automated approaches
based on NLP and rule mining.}.  Newly acquired
constraints are included into the learning problem as additional features. We
extend the regret bounds of \cited{shivaswamy2015coactive} to the more general
case of growing feature spaces. Our empirical findings highlight the promise of
Coactive Critiquing in a synthetic and a realistic preference elicitation
problem, highlighting its ability in offering a reasonable trade-off between
the quality of the recommendations and the cognitive effort expected from the
user.

In the next section we position our work within the related literature. In the
Method section we motivate, detail and analyze our proposed method. We describe
our empirical findings in the Empirical Evaluation section, and conclude with
some final remarks.

\section{Related Work}

There is a large body of work on preference
elicitation~\cite{goldsmith2009preference}. Due to space restrictions, we focus
on the techniques that are most closely related to our approach.

Coactive Learning (\cl) is an interaction model for learning user preferences
from observable behavior~\cite{shivaswamy2012online}, recently employed in
learning to rank and online structured prediction
tasks~\cite{shivaswamy2015coactive,sokolov2015coactive}. For an overview of the
method, see the next section. The underlying weight learning procedure can
range from a simple perceptron~\cite{rosenblatt1958perceptron} to more
specialized online learners~\cite{shivaswamy2015coactive}. Further extensions
include support for approximate inference~\cite{goetschalckx2014coactive} and multi-task
learning~\cite{goetschalckx2015multitask}. These extensions are orthogonal to
our main contribution, and may prove useful when used in tandem. However, in
this paper, we only consider the original formulation, for simplicity.  Our
approach inherits several perks from \cl, including a theoretical
characterization of the average regret~\cite{shivaswamy2015coactive} and native
support for constructive tasks. The main difference between the two methods,
which is also our main contribution, is that in \cc\ the feature space grows
dynamically through critiquing interaction. \cl\ instead works with a static
feature space, and is therefore incapable of handling users with varying
preference criteria.

The concept of critiquing interaction originated in interactive recommender and
decision support
systems~\cite{chen2012critiquing,mcginty2011evolution,tou1982rabbit}.
Critiquing systems invite the user to critique the suggested configurations,
thus supporting the exploration and understanding of the decision domain.
Collected critiques play the role of constraints (or penalties) in filtering
the available options, allowing the search to focus on the more promising
candidates. Our approach is most closely related to user-initiated critiquing
protocols, where at each iteration the user articulates one or more 
critiques~\cite{chen2012critiquing}. In \cc\ critiques are elicited at
specific iterations only, selected by a heuristic balancing cognitive cost and
expressivity of the acquired feature space (as discussed in the Methods
section). Few critiquing recommenders model the user preferences numerically.
In contrast, \cc\ associates weights to both basic and acquired features (i.e.
critiques). One exception is the method of \cited{zhang2006comparative}, which
employs a learned linear utility model.  The user chooses an option from a pool
of $5$ highest utility options.  In this context, critiques are simple
textual descriptions of the advantages of each suggestion over the reference
option.  The estimated utility is updated through a multiplicative update based
on the user's pick. \cc\ instead uses the (user-initiated) critiques to improve
the expressivity of the feature space. Other critiquing recommenders that
include an adaptive component are concerned with developing effective query
selection strategies, e.g. \cite{viappiani2007conversational} and
\cite{viappiani2009regret}.

\section{Method}

We first introduce some notation. We indicate column vectors $\va$ in bold and
vector concatenation as $\va \circ \vb$. The usual dot product is denoted 
$\langle \va, \vb \rangle = \sum_i a_i b_i$ and the Euclidean norm as $\| \va \|$.
Later on we will compute dot products between vectors of different lengths. In
this case, the shorter vector is implicitly zero padded to the bottom to match
the length of the longer one.

\paragraph{Coactive Learning.}
We consider a preference learning setting with coactive feedback; $\calX$ is
the set of feasible item configurations $x$; these are represented by an
$m$-dimensional feature vector $\vphi^*(x)$. We assume that the feature vector
length is bounded, $\|\vphi^*(x)\| \le R$ for some constant $R$. The
attractiveness, or subjective quality, of a configuration is measured by its
\textit{utility}, which we assume~\cite{keeney1976} to be expressible as a
linear function of the features
$u^*(x) := \langle \vw^*, \vphi^*(x) \rangle = \sum_{i=1}^m w^*_i, \phi^*_i(x)$.
Here $\vw^* \in \bbR^m$ encodes the true, unobserved user preferences. We write $x^*$ to
indicate a maximal utility configuration. The goal of the system is to suggest
high utility configurations without direct access to $\vw^*$. A common strategy
is to iteratively improve an estimate of the true preferences through
interaction with the user, while keeping the user's cognitive cost at a
minimum.

We follow the Coactive Learning~\cite{shivaswamy2015coactive} paradigm, which
we describe briefly\footnote{We only consider a ``context-less''
version of Coactive Learning, which is sufficient for our purposes; our method
can be trivially extended to support contexts.
See \cited{shivaswamy2015coactive} for further details.}. In Coactive Learning,
the learner maintains an estimate $\vw^t$ of the user preferences. At each
iteration $t = 1, \ldots, T$, the algorithm computes a most preferable
configuration $x^t \in \calX$, by maximizing the current estimate of the
utility $\langle \vw^t, \vphi^*(x^t) \rangle$. The configuration is then
presented to the user, who is tasked with providing an improved configuration
$\bar{x}^t$, e.g.  by direct manipulation of $x^t$.  The two options $x^t$ and
$\bar{x}^t$ provide an implicit ranking constraint $u^*(\bar{x}^t) > u^*(x^t)$.
The latter is employed to update the preference estimate, in the simplest case
with a perceptron update\footnote{Other update strategies can be applied, see
for instance~\cite{shivaswamy2015coactive}; we will stick with the classical
perceptron with unit step size for simplicity.}:
$$ \vw^{t+1} \gets \vw^t + \vphi^*(\bar{x}^t) - \vphi^*(x^t) $$
In the remainder we assume the user to be
\textit{$\alpha$-informative}~\cite{shivaswamy2015coactive}: if the
configuration $x^t$ is not optimal, the user can always produce an improvement
$\bar{x}^t$ with higher true utility (modulo mistakes). Formally,
$\alpha$-informativity implies that there exists a constant $\alpha \in (0, 1]$
such that, for all $t$, it holds that:
\[ u^*(\bar{x}^t) - u^*(x^t) = \alpha \left( u^*(x^*) - u^*(x^t) \right) - \xi^t \label{eq:alphainformativity} \]
Improvement errors are absorbed by the (possibly negative) slack term $\xi^t
\in \bbR$. Under this assumption, the \textit{average regret} incurred by
Coactive Learning after $T$ iterations, defined as:
\[ REG_T := \frac{1}{T} \sum_{t=1}^T \left( u^*(x^*) - u^*(x^t) \right) \label{eq:avgregret} \]
is bounded from above as follows.

\begin{thm}[\citeauthor{shivaswamy2015coactive}~\citeyear{shivaswamy2015coactive}] \label{thm:cl} For an
$\alpha$-informative user with true preference vector $\vw^*$ and bounded
length feature vectors $\| \vphi^*(x) \| \le R$, the average regret incurred by
Coactive Learning after $T$ iterations is upper bounded by
$$ REG_T \le \frac{2R}{\alpha \sqrt{T}} \| \vw^* \| + \frac{1}{\alpha T}\sum_{t=1}^T \xi^t $$
\end{thm}

As a consequence, so long as the user is not too noisy, the slacks will be
small enough, and the bound guarantees that the average regret will shrink
accordingly.  While similar bounds have been proposed for more general
users~\cite{shivaswamy2015coactive}, here we restrict ourselves to
$\alpha$-informative users for simplicity. In our presentation we do not impose
any restriction on the type of features used. We note in passing, however, that
the choice of feature type can heavily impact the complexity of the inference
step. There are however ways to make Coactive Learning work with approximate
inference procedures~\cite{goetschalckx2014coactive}.

\paragraph{Coactive Critiquing.}
Coactive Learning presupposes the user and the learner to have unlimited access
to the complete feature function $\vphi^*(x)$ at all times. This assumption is
often unrealistic. It is well known that users may discover their own quality
criteria while exploring the option catalogue~\cite{payne1993adaptive};
further, critique queries can be employed to stimulate the users to discover
their own criteria~\cite{faltings2004designing}. We amend to this deficiency by
augmenting Coactive Learning with support for example critiquing interaction.

\begin{algorithm}[tb]
\caption{\label{alg:cc} Pseudo-code of the Coactive Critiquing algorithm. Here
$\vphi^1$ is the initial feature space, and $T$ is the maximum number of
iterations. User interaction occurs inside the \textsc{QueryImprovement} and
\textsc{QueryCritique} procedures.}
\begin{small}
\begin{algorithmic}[1]
    \Procedure{CC}{$\vphi^1,T$}
        \State $\vw^1 \gets 0, \;\; \calD \gets \emptyset$
        \For{$t = 1, \ldots, T$}
            \State $x^t \gets \argmax_{x\in\calX} \langle \vw^t, \vphi^t(x) \rangle$ \label{eq:inference}
            \State $\bar{x}^t \gets \textsc{QueryImprovement}(x^t)$ \label{eq:improvementoracle}
            \State $\calD \gets \calD \cup \{(x^t, \bar{x}^t)\}$
            \If{$\textsc{NeedCritique}(\calD, \vphi^t)$ \label{eq:choiceoracle}}
                \State $\rho \gets \textsc{QueryCritique}(x^t, \bar{x}^t)$ \label{eq:critiqueoracle}
                \State $\vphi^t \gets \vphi^t \circ [\rho] \label{eq:phiupdate}$
                \State $\vw^t \gets \vw^t \circ [0] \label{eq:wpadding}$
            \EndIf
            \State $\vw^{t+1} \gets \vw^t + \vphi^t(\bar{x}^t) - \vphi^t(x^t) \label{eq:update}$
            \State $\vphi^{t+1} \gets \vphi^t$
        \EndFor
        \State $\textbf{return}\; \argmax_{x\in\calX} \langle \vw^T, \vphi^T(x) \rangle$
    \EndProcedure
\end{algorithmic}
\end{small}
\end{algorithm}

At a high level, Coactive Critiquing works as shown in Algorithm~\ref{alg:cc}.
The algorithm maintains estimates of both the user preferences $\vw^t$ and
feature function $\vphi^t(x)$. The initial set of features $\vphi^1(x)$ is
supposedly taken from a reasonable default set, provided by a domain expert,
by the user herself (e.g. through a questionnaire), or other
sources~\cite{chen2012critiquing}. At each iteration $t$, the algorithm
performs an improvement query, as in Coactive Learning
(lines~\ref{eq:inference}, \ref{eq:improvementoracle}), but can additionally
submit a \textit{critique} query to the user. Critiques are only queried when
specific conditions are met (line~\ref{eq:choiceoracle}), as described in the
next subsection.

Given the proposed and improved configurations, $x^t$ and $\bar{x}^t$
respectively, a query critique (line~\ref{eq:critiqueoracle}) amounts to asking
the user why she thinks the improved configuration is preferable to the
suggested one. Ideally, the user would respond with a critique $\rho$ that
maximally explains the utility difference between the two configurations. This
interaction protocol is based on a modest ``local rationality'' assumption:
when presented with two distinct configurations $x^t,\bar{x}^t\in\calX$, the
user can state at least \textit{one} critique that contributes a significant
utility difference between the configurations. The user is free to reply with
suboptimal critiques, according to her current awareness and the required
cognitive effort. We will discuss the impact of suboptimal critiques in our
theoretical analysis.

The feedback of the critique query consists of a single, arbitrary critique
constraint $\rho$. We interpret the latter as a feature function $\rho(x)$ that
captures whether (or how much) the constraint is satisfied. In principle, all
kinds of features are acceptable, including indicators and numerical degrees of
satisfaction. For instance, the critique ``I prefer indoor activities during
winter'' would equate to a feature that indicates the conjunction of the season
being winter and whether the trip includes one or more indoor activities. The
feature $\rho(x)$ is appended to the current feature vector $\vphi^t(x)$; the
weight vector $\vw^t$ is padded accordingly by appending a zero element
(lines~\ref{eq:phiupdate} and \ref{eq:wpadding}). The learner traverses
increasingly more expressive feature spaces $\vphi^t$, $t = 1, \ldots, T$, as
critiques are collected. The perceptron update remains the same as in coactive
learning (line~\ref{eq:update}). The algorithm terminates after a fixed number
of iterations $T$, or when the user is satisfied (e.g. when the regret of the
current suggestion $x^t$ is small enough).

\paragraph{When to ask for critiques.}
Critique queries are key in improving the expressiveness of the feature
space. Critiques are only elicited at the iterations selected by the
\textsc{NeedCritique} procedure (line~\ref{eq:choiceoracle}). The design
of this procedure is crucial. On one hand, if the procedure is too lazy, not
enough critiques are elicited, impairing the representation ability of the
traversed feature spaces. This may in turn make it impossible to learn the true
utility $u^*(x)$. On the other hand, if the procedure is too eager, the algorithm
may end up eliciting more critiques than necessary, thus wasting cognitive
effort. We will show in the next section that, unsurprisingly, the design
of the procedure affects the regret incurred by the learner.

In order to balance between the two, we design a simple selection criterion, as
follows. The idea is to submit a critique query as soon as algorithm realizes
the true utility can no longer be represented in the current feature space.
Since we do not have access to the true utility, we use the collected ranking
feedback (i.e. the dataset, indicated as $\calD$ in Algorithm~\ref{alg:cc}) as
a proxy. To decide whether to ask for a critique or not, we check for the
existence of a weight vector $\vw$ that correctly ranks the pairwise preference
examples in $\calD$, i.e. more formally: $\exists \vw \forall (x,
\bar{x})\!\in\!\calD \ \langle \vw, \vphi(\bar{x}^t)\!-\!\vphi(x^t) \rangle >
0$.

This criterion is guaranteed to work in noiseless scenarios. When the user is
noisy though, a vector $\vw$ satisfying the ranking constraints may not exist
in any subspace of  $\vphi^*(x)$. In this case, Coactive Critiquing may end up
querying for a critique at every iteration. We did not experience this problem
in practice. We also designed a more sophisticated criterion, based on
estimating the likelihood of inconsistencies in the dataset being due to noise
or lack of features. However, we did not see any improvements using this
strategy in our empirical tests.

\paragraph{Theoretical analysis.}
Theorem~\ref{thm:cl} assumes that the feature space is fixed. In coactive
critiquing, however, this is not the case. Our goal is to extend the theorem to
this more general case.

In Coactive Learning, at each iteration $t$, the utility gain provided by two
configurations $\bar{x}^t$ over $x^t$ is $u^*(\bar{x}^t) - u^*(x^t)$, and is
lower bounded by the $\alpha$-informativity assumption
(Eq~\ref{eq:alphainformativity}). Our algorithm however works in a lower
dimensional feature space than the user's one, and has access to the partial
utility $u^t(x) = \langle \vw^*, \vphi^t(x) \rangle$ only. In the lower
dimensional space, the utility gain amounts to $u^t(\bar{x}^t) - u^t(x^t)$, so
it ``misses out'' on the contribution of the unobserved features. We write
$\eta^t$ to denote the missing part, quantified as:
\begin{eqnarray}
    \eta^t
        & :=    & \textstyle \left(u^*(\bar{x}^t) - u^*(x^t)\right) - \left(u^t(\bar{x}^t) - u^t(x^t)\right) \nonumber \\
        & =     & \textstyle \sum_{i=k^t+1}^m w^*_i \left[ \phi^*_i(\bar{x}^t) - \phi^*_i(x^t) \right] \label{eq:eta}
\end{eqnarray}
where $k^t$ is the number of features acquired up to iteration $t$. Note that
$\eta^t$ can be either positive or negative, depending on whether ignoring the
missing features worsens or improves the utility gain, respectively. The latter
case can occur when the $\vphi^*(\bar{x}^t) - \vphi^*(x^t)$ update is
negatively correlated with $\vw^*$ with respect to the missing features.

We formalize this intuition in the following proposition, which is an
adaptation of Theorem~\ref{thm:cl}.

\begin{prop} \label{thm:cc} For an $\alpha$-informative user with true
preference vector $\vw^*$ and bounded length feature vectors $\|
\vphi^*(x) \| \le R$, the average regret incurred by Coactive Critiquing after
$T$ iterations is upper bounded by
$$ REG_T \le \frac{2R}{\alpha \sqrt{T}} \| \vw^* \| + \frac{1}{\alpha T}\sum_{t=1}^T \left( \xi^t + \eta^t \right) $$
\end{prop}

See the Appendix for the proof. The sum $\sum_{t=1}^T \eta^t$ on the right hand
side depends on the effectiveness of the user's critiques and how often they
are asked, as well as the problem structure. The latter factor is beyond our
control, but the former can be (partially) controlled by properly designing the
interaction with the user. By explicitly asking for the critique $\rho$
contributing the most to the utility gain, we are effectively removing the
largest summand from $\eta^t$ (in practice, the user errors may make it
decrease by a smaller amount). Furthermore, the amount of critiques may
reduce the sum of the $\eta^t$, at the price of additional cognitive
effort for the user. In the next section we will show that our proposed
\textsc{NeedCritique} heuristic offers a good trade-off.


\section{Empirical Evaluation}
\label{sec:experiments}

We evaluate Coactive Critiquing on two preference elicitation tasks.  All
experiments were run on a 2.8 GHz Intel Xeon CPU with 8 cores and 32 GiB of
RAM.  Our implementation makes use of MiniZinc~\cite{nethercote2007minizinc}
with the Gecode backend. The \cc\ source code and the full experimental setup
are available at: \texttt{goo.gl/cTFOFq}.

\begin{figure*}[t]
    \centering
    \begin{tabular}{ccc}
        \includegraphics[height=12em]{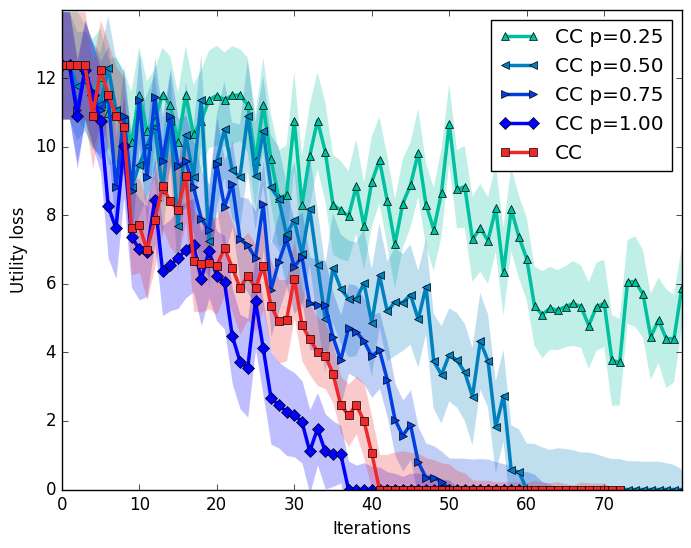} &
        \includegraphics[height=12em]{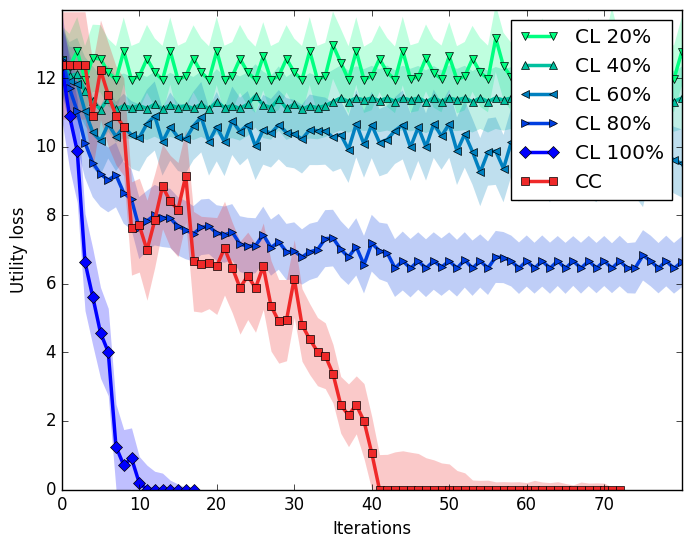} &
        \includegraphics[height=12em]{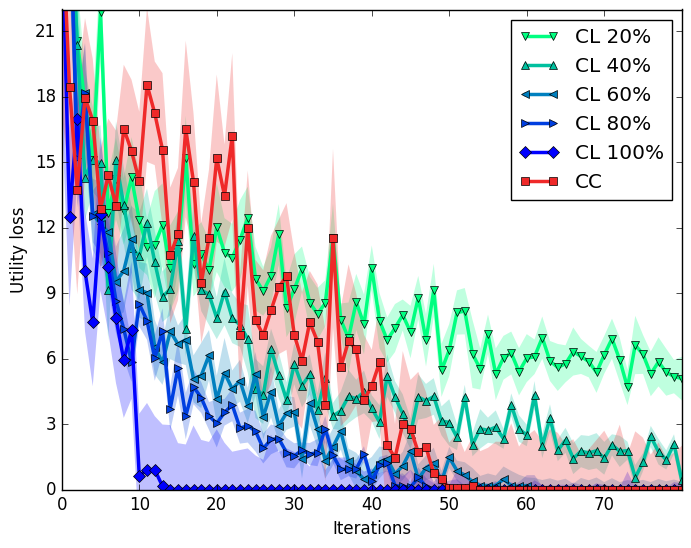}
        \\
        \includegraphics[height=12em]{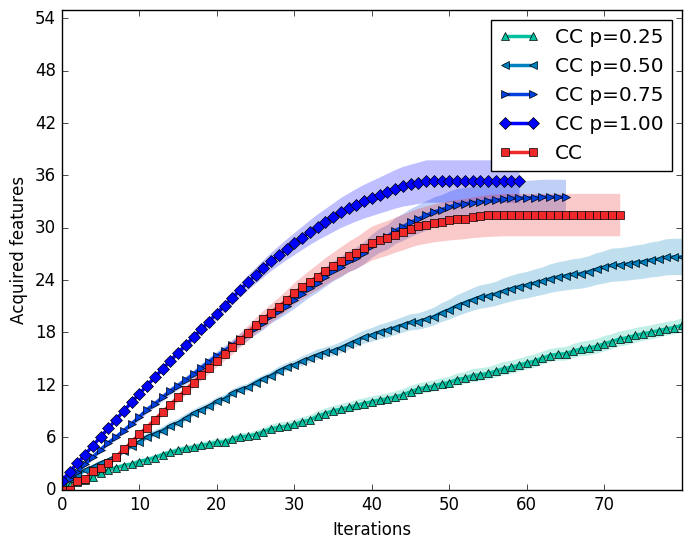} &
        \includegraphics[height=12em]{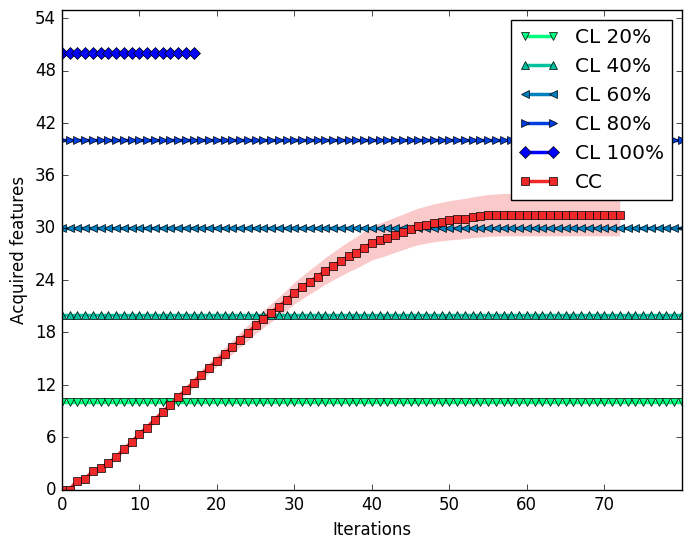} &
        \includegraphics[height=12em]{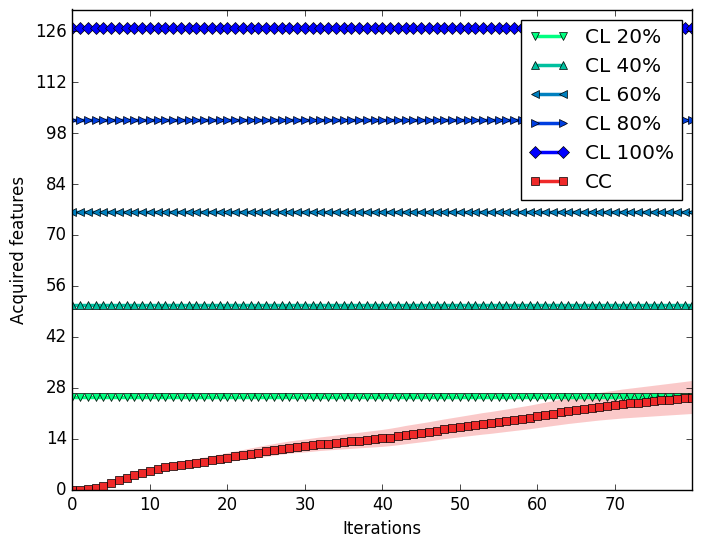}
    \end{tabular}
    \caption{\label{fig:results} Left: comparison of \cc\ for different
choices of \textsc{NeedCritique} procedure; median utility loss at the
top, average number of acquired features at the bottom. Middle: comparison
between \cc\ and \cl\ on the synthetic problem. Right: comparison between \cc\
and \cl\ on the trip planning problem. Best viewed in color.}
\end{figure*}

\paragraph{User simulation.} We simulated the user feedback as follows. In
improvement queries, the user is asked to produce an improvement
$\bar{x}^t$ of the suggested configuration $x^t$. A real user would choose the
improved configuration by balancing between cognitive effort and perceived
quality of the improvement. To account for this fact, our simulated user
(line~\ref{eq:improvementoracle} of Algorithm~\ref{alg:cc}) computes the
improvement by finding a minimal change to $x^t$ with improved true utility.
This is done by solving the combinatorial problem:
\begin{eqnarray*}
    \bar{x}^t := \argmin_{\bar{x} \neq x^t}
        & & \| \bar{x} - x^t \| \\
    \text{s.t.}
        & & \langle \vw^* + \vepsilon, \vphi^*(\bar{x}) \rangle > \langle \vw^* + \vepsilon, \vphi^*(x^t) \rangle
\end{eqnarray*}
Here $\| \bar{x} - x^t \|$ measures the difference between $\bar{x}$ and $x^t$,
and $\vepsilon \in \bbR^m$ is a normally distributed ($\sigma = 0.1$)
perturbation that simulates user noise. The user is $\alpha$-informative as
per Eq~\ref{eq:alphainformativity}. In order not to artificially advantage our
method, our simulated user returns a minimal improvement, consequently providing
a minimal utility gain.

In critiquing queries (line~\ref{eq:critiqueoracle}), the user is asked to
return the critique $\rho$ contributing the most to the utility gain of
$\bar{x}^t$ over $x^t$. Formally, the contribution of feature $\vphi^*_i$ is
$c_i := w^*_i (\phi^*_i(\bar{x}^t) - \phi^*_i(x^t))$. Ideally, the user
would respond with the feature $\rho$
with the highest contribution (with ties broken at random).
In practice, she may choose a sub-optimal critique. We simulate user noise by
sampling $\rho$ from a multinomial distribution where the
probability of choosing $\phi^*_i$ is set to $c_i / \sum_i c_i$. This model
favors features with higher contribution, while still leaving room for
sub-optimal choices.

\paragraph{Synthetic Experiment.} First, we evaluate our method on a synthetic
task. The configurations are 2D points $x$ with integer coordinates, taking
values in a discrete bounding box of size $100 \times 100$, for a total of
$10^4$ feasible configurations. There are $50$ rectangles $r_1, \ldots, r_{50}$
inside the bounding box. The position and size of the rectangles are sampled
uniformly at random once and kept fixed for all runs. Each feature $\phi^*_i(x)$, $i
= 1, \ldots, 50$, acts as an indicator for the corresponding rectangle $r_i$:
it evaluates to $1$ if $x$ is within the rectangle, and to $-1$ otherwise. The
true weights $\vw^* \in \bbR^{50}$ establish a preference over the rectangles:
if $w^*_i > 0$ the user prefers configurations contained in $r_i$, and outside
of it otherwise. It can be readily seen that most features are uncorrelated,
and thus a sufficiently expressive subset of features is needed to find an
optimal solution. The inference and improvement simulators were implemented as
mixed integer-linear problems and solved accordingly.

First, we compare our \textsc{NeedCritique} heuristic against an uninformed
baseline randomly choosing when to ask for critiques. Specifically, we replace
our heuristic at line~\ref{eq:choiceoracle} with a binomial distribution,
varying the parameter $\theta \in \{ 0.25, 0.5, 0.75, 1\}$.

We run all methods over 20 users independently sampled from a $50$-dimensional
standard normal distribution. We compute the median utility loss $u^*(x^*) -
u^*(x^t)$ over all users (the lower, the better) as well as the average number
of acquired features. Execution times are omitted, as the difference between
algorithms is negligible. We report the results in the left column of
Figure~\ref{fig:results}.
As shown by the plots, our heuristic strikes a good balance between user
satisfaction and cognitive effort. In terms of utility loss, it fares
in-between the $\theta=1$ (most informed baseline) and the $\theta=0.75$ (second most
informed) variants, while eliciting fewer critiques than both. The other
baselines are not up to par.

Next, we compare \cc\ with our \textsc{NeedCritique} heuristic against \cl. \cc\
always starts from $2$ features and acquires new ones dynamically through query
critiques. In contrast, \cl\ has fixed access to $p$\% of the features, for
$p \in \{20, 40, \ldots, 100\}$. In order not to bias the results, for each $p$
we take the average of five different \cl\ runs, each over a randomly drawn
subspace of $\vphi^*(x)$ of the appropriate size. We refer to this setting as $\cl^p$. Given that there is no
standard, accepted way to estimate the real cognitive cost of replying to
improvement or critique queries, we avoid computing a single unified measure
of user effort and rather count the number of queries separately. We report the
results in the middle column of Figure~\ref{fig:results}.

In median, $\cl^{100}$ reaches zero loss after 11 iterations, which is hardly
surprising, considering its unrestricted (and unrealistic) access to the full
feature space; $\cc$ instead takes 41 iterations. All other methods fail to
converge. Notably, $\cc$ acquires about 30 features to reach zero median loss,
and beats $\cl^{80}$ in the same metric after 18 iterations, with 14 acquired
features. These results highlight the effectiveness of $\cc$ in acquiring
relevant features, with consequent savings of cognitive effort.

\paragraph{Realistic Experiment.} We applied Coactive Critiquing to an
interactive touristic trip planning task.  We collected a dataset including 10
cities and 15 possible activities from the Trentino Open data website:
\texttt{http://dati.trentino.it/}. The goal is to suggest a trip route $x$
between (some of) the cities. Each city has a particular offering of activities
(e.g.  luxury resorts, points of interest, healthcare services) and an
overnight cost.  Cities may be visited more than once. Traveling between cities
takes a time proportional to their distance. In our experiments we set the trip
length to 10.

We distinguish between base features $\vphi^1(x)$ and full features
$\vphi^*(x)$. The former include the amount of time spent at each location and
the time spent performing each activity, for a total of $25$ base features. The
latter include the number of distinct visited locations, the total time spent
travelling, the total cost, the number of visited geographic regions, among
others, for a total of $92$ acquirable features. We omit the full list for
space restrictions.

As in the synthetic experiment, we compare \cc\ against variants of \cl\
obtained by varying the percentage $p$ of available features over 20 users
sampled from a $127$-dimensional standard normal distribution. We report the
results in the right column of Figure~\ref{fig:results}.

The problem is significantly more difficult than the synthetic one, due to the
combinatorial size of the space of configurations.  The plots show that \cc\ is
very critique-effective: by the last iteration it acquires about as many
features as $\cl^{20}$ (approx. $27$), which is the least informed method, but
it performs considerably better.  The baselines $\cl^{60}-\cl^{100}$ converge
faster, having access to most of the features from the beginning. Our approach
performs comparably to $\cl^{60}$ from iteration 50 onwards, notwithstanding
the much fewer acquired features ($\sim\!\!20$ versus $\sim\!\!75$,
respectively).  Although $\cl^{40}$ uses (from iteration $1$) about twice the
number of features eventually acquired by \cc, it is surpassed by the latter
roughly at the $40^{\text{th}}$ iteration.

\section{Conclusion}
\label{sec:conclusion}

In this paper we described an approach to preference elicitation that combines
Coactive Learning with example critiquing interaction. Contrary to coactive
learning, the feature space is acquired dynamically through interaction with
the user. We discussed the theoretical guarantees of the method, and a
heuristic query selection strategy that balances between user effort and
expressivity of the acquired feature space. We presented experimental evidence
in support of our findings. Coactive Critiquing is competitive with more
informed baselines, often requiring many less features to obtain comparable (or
better) recommendations.  Like conversational recommenders, Coactive Critiquing
could in principle handle free-form textual or speech critiques,
see for instance \cited{grasch2013recomment}.

Coactive Critiquing is especially suited for constructive preference
elicitation tasks~\cite{teso2016constructive}. Given that the computational
cost of inference can become prohibitive in these settings, it may be fruitful
to integrate support for approximate inference, as discussed
by \cited{goetschalckx2014coactive}.
Another promising research direction involves allowing the user to reply with
non-feasible improved configurations. In this case, the projection of the
improvement on the feasible space may break $\alpha$-informativity. We are
currently investigating how to tackle this issue.

\subsubsection{Acknowledgments}
ST is supported by the CARITRO Foundation through grant 2014.0372. PD is a
fellow of TIM-SKIL Trento and is supported by a TIM scholarship.

\appendix
\section{Appendix A: Proof of Proposition~\ref{thm:cc}}

We split the proof in three steps.

(i) The update equation of Algorithm~\ref{alg:cc} (line \ref{eq:update}) is
$$ \vw^{T+1} := \vw^T + \vphi^T(\bar{x}^T) - \vphi^T(x^T) $$
We expand the dot product $\langle \vw^{T+1}, \vw^{T+1} \rangle$ using the
above, obtaining
\begin{eqnarray*}
    & & \textstyle \langle \vw^T, \vw^T \rangle + 2 \langle \vw^T, \vphi^T(\bar{x}^T) - \vphi^T(x^T) \rangle + \\
    & & \textstyle \qquad \langle \vphi^T(\bar{x}^T) - \vphi^T(x^T), \; \vphi^T(\bar{x}^T) - \vphi^T(x^T) \rangle
\end{eqnarray*}
The optimality of $x^T$ in the current feature space $\vphi^T(x)$ (line
\ref{eq:inference}) implies that the second term is no greater than zero.
Given that $\|\vphi^T(x)\| \le \|\vphi^*(x)\| \le R$ by assumption, it follows
that
\[ \textstyle \langle \vw^{T+1}, \vw^{T+1} \rangle \le \langle \vw^T, \vw^T \rangle + 4R^2 \le 4R^2T \label{eq:step1} \]

(ii) Let $\vz^T$ be a 0-1 vector, of the same shape as $\vw^*$ such that the
only non-zero elements of $\vz^T$ are those corresponding to the features
elicited up to iteration $T$.
We expand the dot product $\langle \vw^*, \vw^{T+1} \rangle$ using the above
update rule to obtain
\begin{eqnarray*}
    & & \textstyle \langle \vw^*, \vw^T \rangle + \langle \vw^*, \vphi^T(\bar{x}^T) - \vphi^T(x^T) \rangle = \\
    & & \qquad \textstyle \langle \vw^*, \vw^T \rangle + \langle \vw^*, \vz^T \odot \left[ \vphi^*(\bar{x}^T) - \vphi^*(x^T) \right] \rangle
\end{eqnarray*}
where $\odot$ is the element-wise product. We unroll the recursion to get
\begin{eqnarray*}
    \textstyle \langle \vw^*, \vw^{T+1} \rangle = \sum_{t=1}^T & & \hspace{-1.6em} \textstyle \langle \vw^*, \vz^t \odot \left[ \vphi^*(\bar{x}^t) - \vphi^*(x^t) \right] \rangle \\
    \textstyle = \sum_{t=1}^T & & \hspace{-1.6em} \textstyle \langle \vw^*, \vphi^*(\bar{x}^t) - \vphi^*(x^t) \rangle - \\
    & & \hspace{-1.2em} \textstyle \langle \vw^*, (\mathbf{1} - \vz^t) \odot \left[ \vphi^*(\bar{x}^t) - \vphi^*(x^t) \right] \rangle
\end{eqnarray*}
By applying the definition of utility $u^*(x)$ to the first term and the
definition of $\eta^T$ to the second one, we get
\[ \textstyle \langle \vw^*, \vw^{T+1} \rangle = \sum_{t=1}^T \left[ u^*(\bar{x}^t) - u^*(x^t) \right] - \sum_{t=1}^T \eta^t \label{eq:step2} \]

(iii) The Cauchy-Schwarz inequality states that
$\langle \vw^*, \vw^{T+1} \rangle \le \| \vw^* \| \| \vw^{T+1} \|$.
We plug Equations~\ref{eq:step1} and \ref{eq:step2} to obtain:
$$ \textstyle \sum_{t=1}^T \langle \vw^*, \vphi^*(\bar{x}^t) - \vphi^*(x^t) \rangle \le 2R \sqrt{T} \| \vw^* \| + \sum_{t=1}^T \eta^t $$
Now we use the $\alpha$-informativity assumption
(Eq~\ref{eq:alphainformativity}) and the definition of average regret
(Eq~\ref{eq:avgregret}) to obtain the claim.

\bibliographystyle{aaai}
\bibliography{paper}

\end{document}